\documentclass{nature_meth}
\usepackage{amssymb,amsfonts,amsmath}
\usepackage{graphicx} 
\usepackage{gensymb}

 \usepackage{multirow}
\usepackage{epsfig}
\usepackage[figoff]{figcaps}
\usepackage{hyperref}
\hypersetup{
    colorlinks=true,
    linkcolor=blue,
    filecolor=magenta,
    urlcolor=cyan,
}

\usepackage{outline}
\usepackage{pmgraph}
\usepackage[normalem]{ulem}
\usepackage[utf8]{inputenc}
\usepackage{amssymb}
\usepackage{hyperref}
\usepackage{amsmath}
\usepackage{graphicx}
\usepackage{times}
\usepackage{xcolor}
\usepackage{xspace}
\usepackage[colorinlistoftodos]{todonotes} 
\usepackage{cite}
\usepackage{bm} 
\usepackage{url}

\usepackage{epstopdf}
\AppendGraphicsExtensions{.tiff}
\graphicspath{{fig/}} 

\usepackage{epsfig}
\usepackage{tikz}
\usetikzlibrary{spy}
\usepackage{algpseudocode}
\usepackage{algorithm}
\usepackage{mathrsfs}

\long\def\comment#1{} 

\newcommand{\xmath}[1] {\ensuremath{#1}\xspace}
\newcommand{\blmath}[1] {\xmath{\bm{#1}}}

\newcommand{\Xb}{{\blmath X}}

\newcommand{\cb}{{\blmath c}}

\newcommand{\xb}{{\blmath x}}
\newcommand{\yb}{{\blmath y}}
\newcommand{\zb}{{\blmath z}}

\newcommand{\Ac}{\mathcal{A}}

\newcommand{\Xc}{\mathcal{X}}
\newcommand{\Yc}{\mathcal{Y}}

\newcommand{\Rd}{{\mathbb R}}

\newcommand{\beq}{\begin{equation}}
\newcommand{\eeq}{\end{equation}}
\newcommand{\beqa}{\begin{eqnarray}}
\newcommand{\eeqa}{\end{eqnarray}}

\usepackage{caption,setspace}
\renewcommand{\spacing}[1]{\renewcommand{\baselinestretch}{#1}\large\normalsize}
\spacing{2}

\usepackage{graphicx}
\makeatletter
\let\saved@includegraphics\includegraphics
\AtBeginDocument{\let\includegraphics\saved@includegraphics}
\renewenvironment*{figure}{\@float{figure}}{\end@float}
\makeatother

\title{Feature Disentanglement in generating three-dimensional structure from two-dimensional slice with sliceGAN}
\author{Hyungjin Chung$^{1}$ and Jong Chul Ye$^{\dagger,1}$
}

\begin{document}
\maketitle

\begin{affiliations}
\item Department of Bio and Brain Engineering, KAIST, Daejeon, Korea
\item[] $^{\dagger}$Correspondence should be addressed to J.C.Y. (jong.ye@kaist.ac.kr)
\end{affiliations}
 

Deep generative models are known to be able to model arbitrary probability distributions. Among these, a recent  deep generative model, dubbed sliceGAN~\cite{kench2021generating}, proposed a new way of using the
generative adversarial network (GAN)  to capture the micro-structural characteristics of a two-dimensional (2D) slice and generate three-dimensional (3D) volumes with similar properties. While 3D micrographs are largely beneficial in simulating diverse material behavior, they are often much harder to obtain than their 2D counterparts. Hence, sliceGAN opens up many interesting directions of research by learning the representative distribution from 2D slices, and transferring the learned knowledge to generate arbitrary 3D volumes. 

In the original implementation of sliceGAN, the authors proposed training a single 3D generator which takes in as input a noise vector sampled from a uniform distribution, as well as  three separate discriminators that enforce the probability distributions of the generated slices in $x, y,$ and $z$ dimensions to be realistic.  At every optimization step, the discriminator takes in every 2D slice from the generated 3D volume as {\em fake} data, while the same number of random patches from the real 2D slice image are fed as {\em real} data. As the training progresses, the generator produces more and more realistic 2D slices in every dimensions, and consequently, once the training is complete, the generator is able to synthesize 3D volumes having the same characteristics of the original 2D slice. 

\begin{figure}[!b]
	\centering
 \centerline{\epsfig{figure=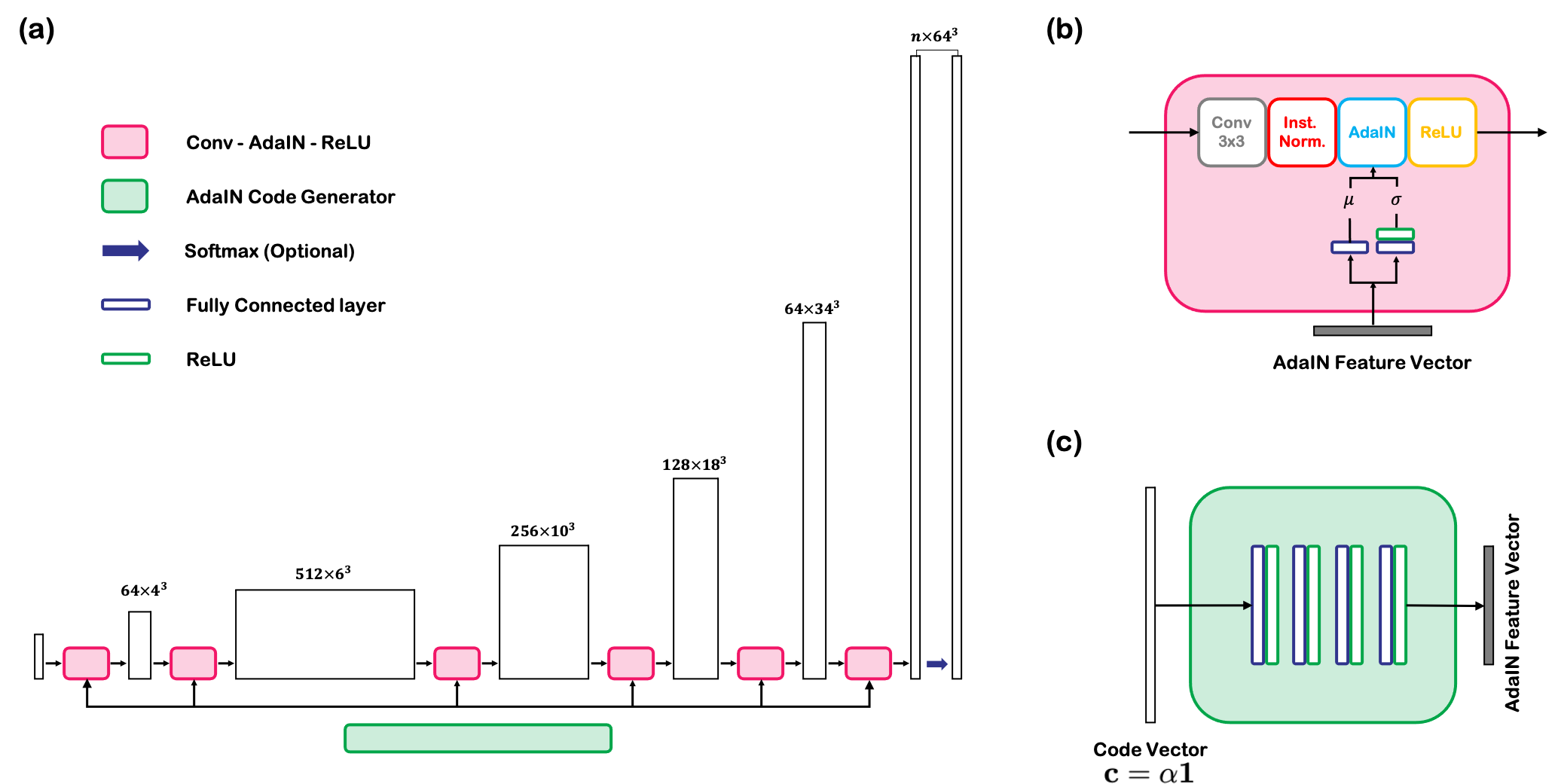, width=1.0\linewidth}}
	\caption{\bf\footnotesize 
		Overview of AdaIN-style generator for continuous feature disentanglement of generated three-dimensional volume. (a) Generator architecture, (b) basic building block of the generator, and (c) AdaIN code generator.}
	\label{fig:overall_flow}
\end{figure}

Notably, sliceGAN mostly focuses on the generation of $n$-phase materials, which are implemented using $n$ different channel output and the final softmax layer as shown in Figure~\ref{fig:overall_flow}(a). It is trivial that the discrete modeling used in $n$-phase material modelling is easier than estimating the probability distribution of a continuous data, since in the discrete case the voxel only belongs to one of $n$ classes. Consistent with the authors' claim~\cite{kench2021generating}, when we tried reproducing the training step using both $n$-phase data and greyscale data, we acquired considerably better results when modelling $n$-phase data.

Another limitation of the prior work~\cite{kench2021generating} is that we cannot control the characteristic of the generated samples at inference phase. Any noise vector sampled from uniform random distribution is capable of generating 3D volumes of the same characteristics, but latent space steering is not possible. Consequently, even when different data share the same features except for one feature such as grain size, sliceGAN needs to be trained for every specific datum, which takes about 4 hours per single datum. 

On the other hand,  over the several years of advances {in} GAN, it has now become common to be able to control the visual attributes of the sample being generated from the input noise vector  by steering feature vectors. One of the most famous and widely used examples is styleGAN~\cite{karras2019style}, where adaptive instance normalization (AdaIN)~\cite{huang2017arbitrary} is adopted to inject style at certain layers of the generator. This so-called style module supposedly disentangles the feature space, and by constructing the network this way, each layer is able to control different attributes of the data being generated. In the context of medical imaging~\cite{yang2020continuous, gu2021adain}, AdaIN was combined with cycleGAN~\cite{zhu2017unpaired} for continuous conversion between two domains $\Xc$, and $\Yc$. Leveraging the idea of feature disentanglement, in this reusability report, we present a simple extension of sliceGAN by endowing the model the ability to control the attributes of the generated 3D volume. By this extension, sliceGAN can be freed from having to be trained per every specific dataset, but can be trained with different data from the same class. Then, once the network is trained, controlled synthesis of diverse data in the same class is possible.

\section*{Feature disentanglement in SliceGAN using AdaIN}

Consider the feature tensor at a specific layer:
\begin{equation}
    \Xb = [\xb_1 \dots \xb_C] \in \Rd^{HWD \times C},
\end{equation}
where $\xb_i$ is the $i$-th channel feature vector of $\Xb$ containing a single-channel volume feature of size $\Rd^{H \times W \times D}$. To the volume feature, we can utilize AdaIN to inject some arbitrary information $\yb$ at each channel dimension, formally described as
\begin{equation}
    \zb_i = \Ac(\xb_i, \yb_i), \quad i = 1, \dots, C,
\end{equation}
with
\begin{equation}
    \Ac(\xb, \yb) := \frac{\sigma(\yb)}{\sigma(\xb)}(\xb - \mu(\xb)\mathbf{1}) + \mu(\yb)\mathbf{1}.
\end{equation}
Here, $\mu(\cdot)$ and $\sigma(\cdot)$ refers to mean and standard deviation, respectively, and $\mathbf{1}$ is a vector filled with ones of size $\Rd^{HWD}$. 
In our extension of sliceGAN, this AdaIN transform is performed after every convolution of the sliceGAN generator as shown in Figure~\ref{fig:overall_flow}(a)(b). Now, to inject meaningful feature to $\yb$, we utilize AdaIN code generator shown in Figure~\ref{fig:overall_flow}(c), which is a simple MLP with ReLU nonlinearity, taking in a static vector representing certain features such as grain size, and outputs a AdaIN feature vector to be used as $\yb$. Once the network is trained with AdaIN-style generator, we can simply change the code vector at the inference phase while keeping the same noise vector.
 
Specifically, to control feature that represents  the grain size,  
in our implementation the code vector is defined as $\cb=\alpha\mathbf{1}$, where $\alpha$ is a scale parameter and $\mathbf{1}$ is a vector filled with ones.
Then, it is remarkable that feature interpolation is possible at inference phase~\cite{yang2020continuous}. For example, in the generation
of code vector, if a small scale parameter $\alpha$ was used together with data of small grain size and a larger scale parameter $\alpha'$ was used together with data of large grain size, then when we inject {a} value in between $\alpha$ and $\alpha'$ to interpolate the code vector, the generator synthesizes a volume with medium grain size, which we further elaborate with empirical results in the following section.

\section*{Experiments and Implementation}

The data that we chose as our proof-of-concept was acquired from an open source database~\cite{hsu2018mesoscale}. The database contains microstructures of three phase synthetically generated with DREAM3D~\cite{groeber2014dream}, with log-normal feature size distribution for each phase. Three structures with size 75 $\times$ 75 $\times$ 15 $\mu$m$^3$, 125 nm voxel size, mean feature diameter of 0.57 $\mu$m$^3$ were acquired with varying scale parameter. More specifically, the scale parameter was represented as the log of geometric standard deviation, $\ln \sigma_g$, and the value for each data was specified to 0.15, 0.35, and 0.6. Consequently, the input code vector $\cb$ represented in Figure~\ref{fig:overall_flow}(c) has the shape $128 \times 1$, with every element set to the same scale parameter value $\alpha \in \{0.15, 0.35, 0.6\}$ for each target data.

For implementation, we make minimal changes to the codebase provided by the authors of sliceGAN~\cite{kench2021generating}. We keep the generator and {the} discriminator architecture the same as in the original paper, only replacing the batch normalization block with {the} AdaIN block as shown in Figure~\ref{fig:overall_flow}(a-c). Training strategies, including the choice of generator/discriminator update steps and respective batch size were deliberately kept identical to the original work, and we found that it was sufficient to produce high-quality results.

\begin{figure}[!hbt]
	\centering
 \centerline{\epsfig{figure=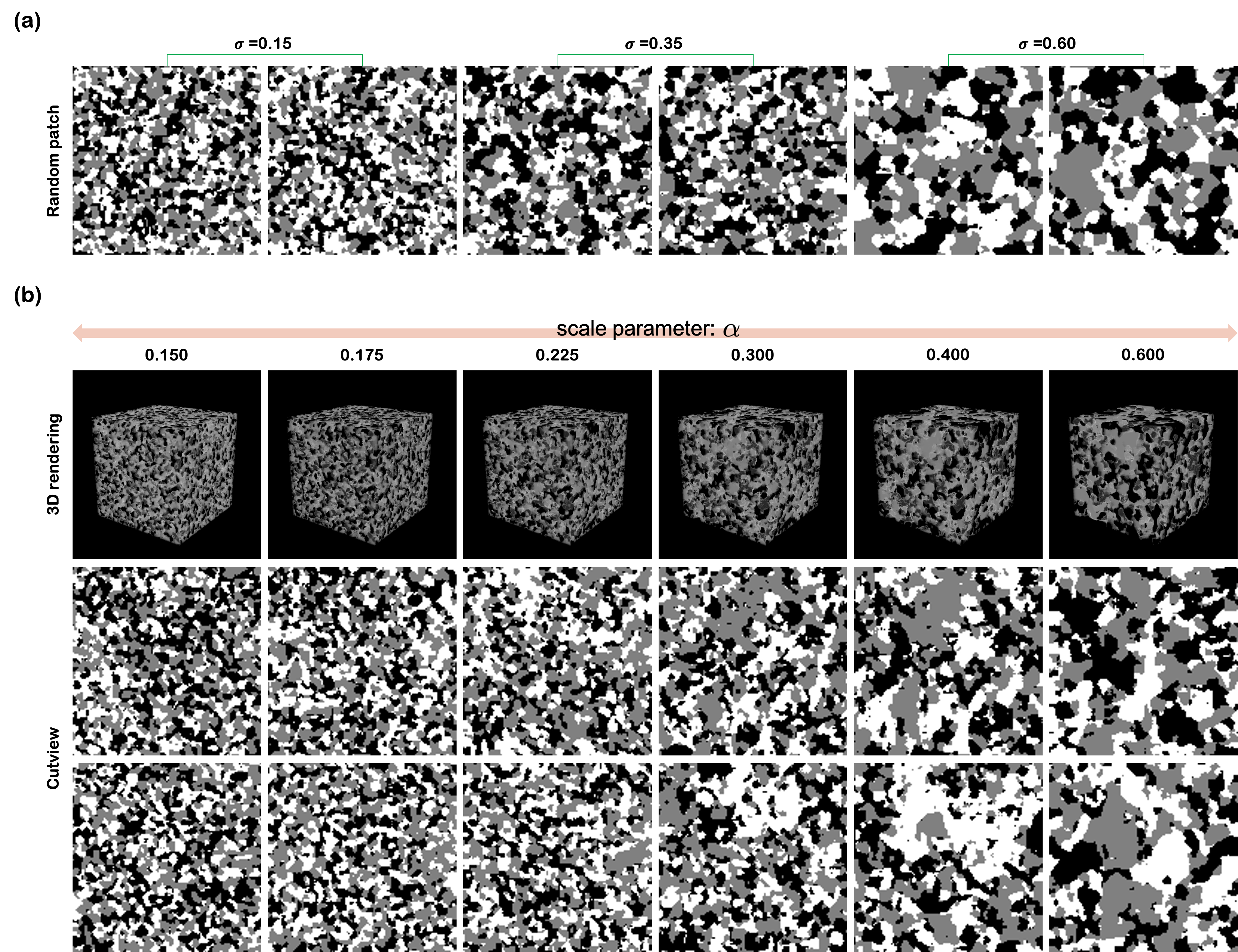, width=0.9\linewidth}}
	\caption{\bf\footnotesize 
		(a) Random patches taken from the source target images. (b) Synthesized results by  the code vector $\cb=\alpha \mathbf{1}$ from the interpolated scale parameter $\alpha$. All results are generated from a single generator.  The first row shows the 3D rendered view of each data, and the second and third rows show $x-y$, $y-z$ plane cutviews, respectively.}
	\label{fig:results}
\end{figure}

In Figure~\ref{fig:results}, we summarize the results. Although there are only three different values of code vector used at the training stage as presented in Figure~\ref{fig:results}(a), when we select any values within the range, the generator synthesizes feasible results, as shown in Figure~\ref{fig:results}(b). Hence, now that the feature is disentangled, we can choose any values of scale parameter $\alpha$ according to the desired grain size. It should also be denoted that we also tried extrapolation outside the range of the training set. However, values smaller than 0.15 did not synthesize smaller grain, and values larger than 0.60 did not synthesize larger grain.

\section*{Discussion}

SliceGAN~\cite{kench2021generating} was verified to be well-applicable to synthesis of various $n$-phase materials. Without any fine-tuning, we could acquire satisfactory results as presented in Figure~\ref{fig:results}, showing the robustness of the work. Our extension of disentangling the feature so that it is controllable at inference phase was straightforward, and the extension did not require any change from the original work on medical imaging~\cite{yang2020continuous,gu2021adain}. We believe that our approach would be applicable to {a} different dataset with features other than the grain size, which requires a curated dataset.

The training data acquired from \cite{hsu2018mesoscale} were not strictly a single slice of 2D slice, but of multiple slices of large matrix size. We tried both training sliceGAN from random patches of single slice, and training sliceGAN from random patches cut from multiple slices, and found that both approaches produce stable results without mode collapse. Nonetheless, we conjecture that the model will benefit from using multiple slices for training data, but this effect is not clearly observable in the case of simple $n$-phase materials that are successfully trained from even a single slice.

\section*{Future directions}

First, extending the work of sliceGAN to the generation of continuous greyscale microstructures is a promising direction of research. This can be carried out along many directions. For example, the generator and discriminator architectures used in sliceGAN are simplistic, having a single convolutional block per each scale. Consequently, we believe that scaling the architecture to a larger size would be a benefit, although it would require a larger memory GPU (Original work~\cite{kench2021generating} used 12GB GPU, and we used 11GB GPU). Moreover, progressive learning strategy as in \cite{karras2017progressive, karras2019style} could also strengthen the method.
Once the generation of greyscale microstructures become robust, we believe that the synthesis of RGB color volumes will also work easily, by simply expanding the channel dimension.

 \begin{addendum}
{\color{black} \item[Correspondence] Correspondence and requests for materials should be addressed to Jong Chul Ye.~(email: jong.ye@kaist.ac.kr).}
 \item  This research was funded by the National Research Foundation (NRF) of Korea grant NRF-2020R1A2B5B03001980.
{
\item[Author Contributions] H.C. performed all experiments, wrote the extended code, and prepared the manuscript. J.C.Y. supervised the project in conception and discussion, and prepared the manuscript.}
 \item[Competing Interests] 
The authors declare that they have no competing financial interests.
 {
  \item[Data Availability] 
The data is accessible at ref.\cite{hsu2018mesoscale}
  }
  {
  \item[Code Availability] 
The code is available at the following \href{https://github.com/bispl-kaist/SliceGAN_AdaIN}{link}
  }
\end{addendum}

\bibliographystyle{naturemag}
\bibliography{ref}

\end{document}